\documentclass[letterpaper]{article} 
\usepackage{aaai2026}  
\usepackage{times}  
\usepackage{helvet}  
\usepackage{courier}  
\usepackage[hyphens]{url}  
\usepackage{graphicx} 
\usepackage{natbib}  
\usepackage{caption} 
\usepackage{algorithm}
\usepackage{algorithmic}

\usepackage{graphicx}%
\usepackage{pifont}%
\usepackage{multirow}%
\usepackage{amsmath,amssymb,amsfonts}%
\usepackage{amsthm}%
\usepackage{mathrsfs}%
\usepackage{multirow}
\usepackage{float}
\usepackage{subfloat}
\usepackage{booktabs}%
\usepackage{caption}
\usepackage{color}

\usepackage[capitalize]{cleveref}
\usepackage{newfloat}
\usepackage{listings}
\DeclareCaptionStyle{ruled}{labelfont=normalfont,labelsep=colon,strut=off} 
\floatstyle{ruled}
\newfloat{listing}{tb}{lst}{}
\floatname{listing}{Listing}
\title{Enhancing Noise Resilience in Face Clustering via Sparse Differential Transformer}
\author{
    Dafeng Zhang, Yongqi Song, Shizhuo Liu
}
\affiliations{
    Samsung R\&D Institute China-Beijing (SRC-B), Beijing, China \\

    {dfeng.zhang, yongqi.song, shizhuo.liu}@samsung.com
}

\usepackage{bibentry}

\begin{document}

\maketitle

\begin{abstract}
The method used to measure relationships between face embeddings plays a crucial role in determining the performance of face clustering. Existing methods employ the Jaccard similarity coefficient instead of the cosine distance to enhance the measurement accuracy. 
However, these methods introduce too many irrelevant nodes, producing Jaccard coefficients with limited discriminative power and adversely affecting clustering performance. To address this issue, we propose a prediction-driven Top-K Jaccard similarity coefficient that enhances the purity of neighboring nodes, thereby improving the reliability of similarity measurements. Nevertheless, accurately predicting the optimal number of neighbors (Top-K) remains challenging, leading to suboptimal clustering results. To overcome this limitation, we develop a Transformer-based prediction model that examines the relationships between the central node and its neighboring nodes near the Top-K to further enhance the reliability of similarity estimation. However, vanilla Transformer, when applied to predict relationships between nodes, often introduces noise due to their overemphasis on irrelevant feature relationships. To address these challenges, we propose a Sparse Differential Transformer (SDT), instead of the vanilla Transformer, to eliminate noise and enhance the model's anti-noise capabilities. Extensive experiments on multiple datasets, such as MS-Celeb-1M, demonstrate that our approach achieves state-of-the-art (SOTA) performance, outperforming existing methods and providing a more robust solution for face clustering.
\end{abstract}

\section{Introduction}
\label{Introduction}

Face clustering is a critical task in computer vision and machine learning, with numerous applications in areas such as photo management, social media organization, and data collection for face recognition models~\cite{guo2016ms, zhu2021webface260m, deng2019arcface}. The rapid growth of digital images, particularly face images, has made it increasingly important to develop efficient and accurate methods for grouping similar faces together. Traditional clustering methods, such as K-Means~\cite{lloyd1982least} and DBSCAN~\cite{ester1996density}, have been widely used for this purpose. However, these methods rely on assumptions about data distribution and are sensitive to hyperparameters, which limits their effectiveness on large-scale and complex datasets.

\begin{figure}[t]
	\centering
	\includegraphics[width=0.99\linewidth]{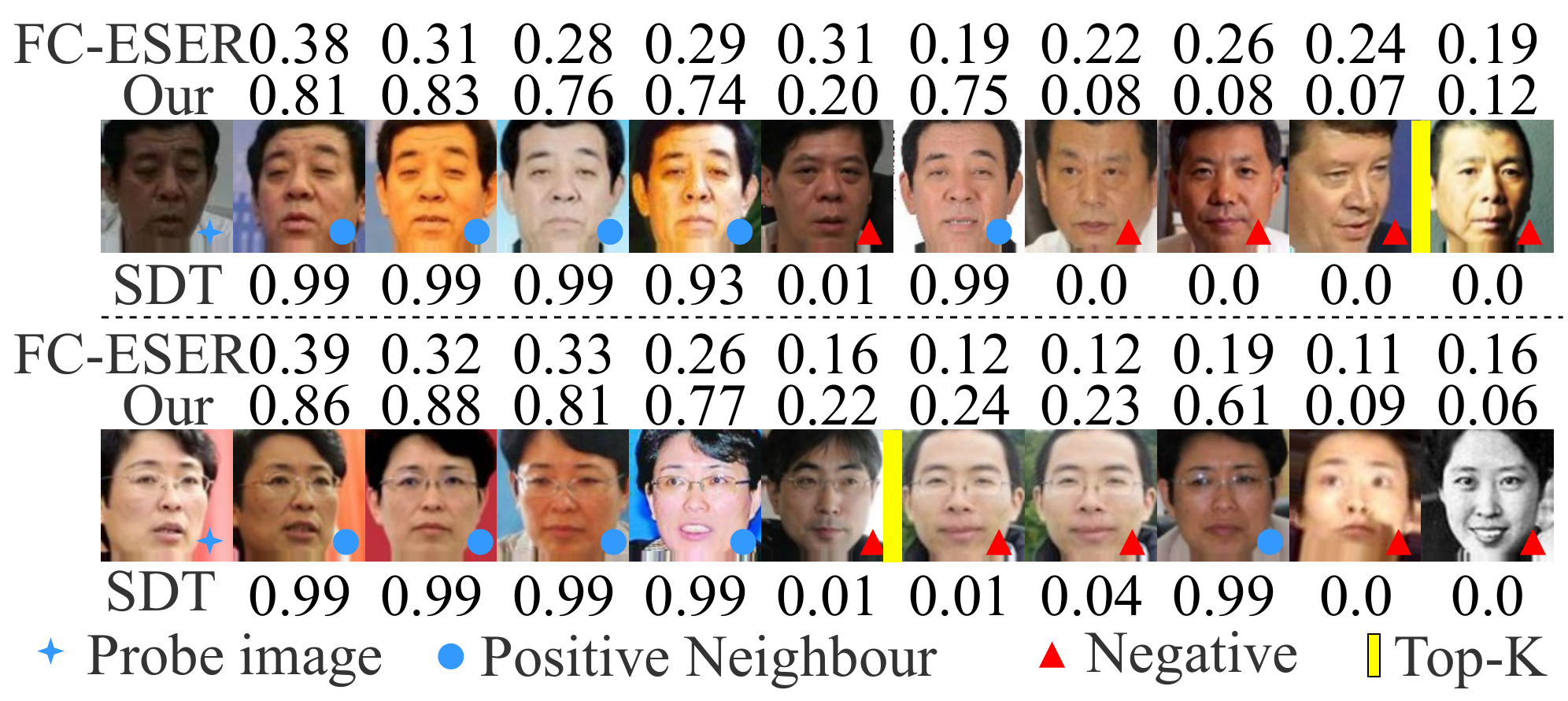}
	\vspace{-2mm}
	\caption{The facial similarity score with different methods. } 
	\label{fig:1}
	\vspace{-3mm}
\end{figure}

Recent years have witnessed remarkable advancements in face clustering, largely driven by the power of Graph Convolutional Networks (GCN)~\cite{kipf2016semi, wang2019linkage, yang2019learning, yang2020learning, shen2021structure, liu2021learn, wang2022ada, shen2023clip}. These networks excel in learning and propagating features, enabling more accurate clustering. However, a significant hurdle persists in the form of noise edges~\cite{wang2022ada}. When constructing face graphs, the traditional approach of relying on $k$NN~\cite{cover1967nearest} relations based on cosine distance in the feature space often leads to the inclusion of edges connecting faces of different identities. 
As a result, when messages propagate along these noise edges during GCN operations, the face features get contaminated, degrading the overall clustering performance.
Ada-NETS~\cite{wang2022ada} and FC-ESER~\cite{liu2024face} represent a significant step forward in addressing this issue. They introduce the Jaccard similarity coefficient~\cite{zhong2017re}, instead of the cosine distance, to measure the similarity between two samples in the face graph. However, they introduce a large number of irrelevant nodes, resulting in highly similar Jaccard coefficients that degrade clustering performance. 
As shown in~\cref{fig:1}, the Jaccard similarity coefficients computed by FC-ESER for different faces are very close, which introduces challenges to discriminative face identification. 
An excessively low threshold risks merging faces of differing identities into a single cluster, whereas an overly high threshold fragments faces of the same identity into multiple clusters.

In order to improve the reliability of Jaccard similarity coefficient, we propose a prediction-driven Top-K Jaccard similarity coefficient in this paper. By carefully and dynamically selecting Top-K true neighboring nodes for each central node, the purity of neighboring nodes considered when calculating Jaccard similarity coefficients can be improved. Conversely, this will make the Jaccard similarity coefficient more reliable, reduce the influence of noisy nodes, and minimize the chance of clustering errors caused by the inclusion of irrelevant nodes.
As shown in~\cref{fig:1}, our method captures facial similarity more accurately than FC-ESER~\cite{liu2024face}.

Another critical issue lies in the instability of the predicted Top-K near its optimal number of neighbors, which may either omit relevant nodes or introduce irrelevant ones. This necessitates a thorough examination of the relationships between the central node and its neighboring nodes near the Top-K. To address this problem, we develop a Transformer-based model to predict these relationships in this paper. However, vanilla Transformer~\cite{vaswani2017attention, nguyen2021clusformer, ye2021learning, chen2022mitigating, wang2022robust, shin2023local, liu2024face} tends to allocate excessive attention to all feature relationships, including those that are irrelevant or noisy~\cite{ye2024differential}. This overemphasis can cause the model to misinterpret relationships, leading to incorrect clustering decisions. 
In response to the noise attention issue associated with the Vanilla Transformer~\cite{vaswani2017attention}, we propose the Sparse Differential Transformer (SDT) to replace the above Transformer-based network. Inspired by the Differential Transformer~\cite{ye2024differential}, which aims to cancel attention noise through differential denoising, our SDT takes it a step further. Specifically, we propose a Top-K sparse differential attention model that only focuses on the related nodes before the Top-K and masks the attention of unrelated nodes when calculating the attention score. This design enables the model to focus more precisely on the most relevant facial features, significantly improving its anti-noise ability and ultimately enhancing the clustering performance.

The contributions can be summarized as follows:
\begin{itemize}
	\item We propose a prediction-driven Top-K Jaccard similarity coefficient that enhances the purity of neighboring nodes, to improve the reliability of similarity measurements.

	\item Our neural network-based approach for predicting neighbor relationships near the Top-K offers a more reliable strategy for adaptive neighbor discovery, leading to better clustering performance.
	
	\item We propose a Sparse Differential Transformer (SDT) network to address the noise attention problem in Vanilla Transformer, improving the model's anti-noise ability.
	
	\item The comprehensive experiments on multiple datasets demonstrate that our method achieves state-of-the-art (SOTA) performance, outperforming existing methods in both clustering accuracy and robustness.
	
\end{itemize}

\section{Related Work}
\label{sec:Related Work}

\begin{figure*}[t]
	\centering
	\includegraphics[width=0.88\linewidth]{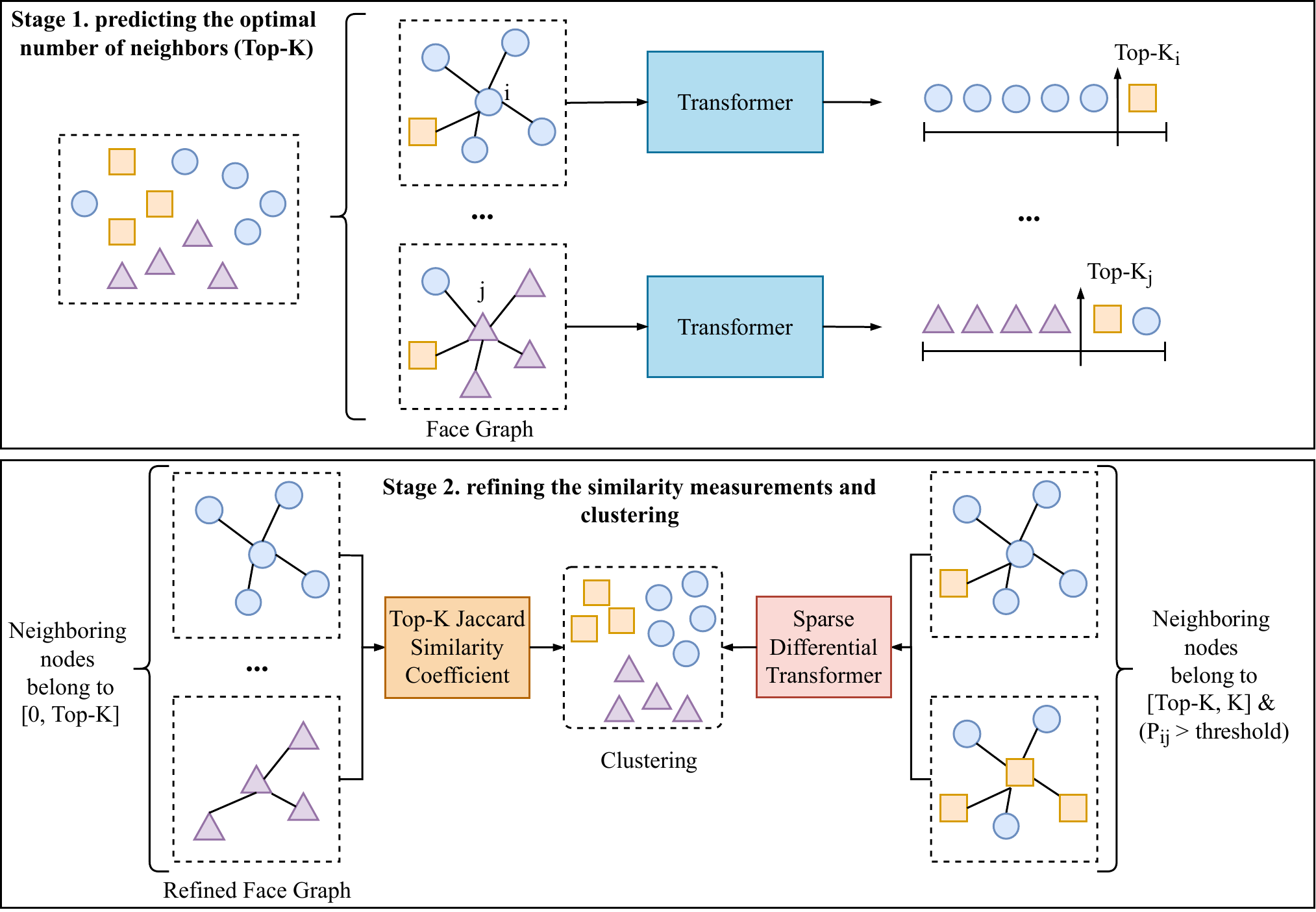}
	\vspace{-1mm}
	\caption{The framework and network architecture of our face clustering method. } 
	\label{fig:4}
	\vspace{-2mm}
\end{figure*}

\subsection{Face Clustering}

Traditional clustering algorithms such as K-Means~\cite{lloyd1982least}, Hierarchical Agglomerative Clustering (HAC)~\cite{sibson1973slink}, and DBSCAN~\cite{ester1996density} have been widely used for face clustering tasks~\cite{ho2003clustering, zhao2006automatic}. These methods operate on the assumption of convex cluster shapes and homogeneous compactness, which are often impractical for the diverse and complex nature of face data. In recent years, GCN-based supervised methods have shown great promise. 
L-GCN~\cite{wang2019linkage} deploys a GCN for linkage prediction on subgraphs, leveraging the graph structure to infer connections between face images. CAGCN~\cite{zhang2023context} enhances the feature expression capabilities by learning and fusing both global and local features. LTC~\cite{yang2019learning} formulates clustering as a detection and segmentation pipeline based on GCN, while GCN(V+E)~\cite{yang2020learning} proposes a confidence estimator and a connectivity estimator to estimate the vertex confidence and edge connectivity in the face graph. They are all two-stage GCN method. 
Despite the significant progress made by these methods, they still face challenges. Most build face graphs based on $k$NN, which often results in a significant number of noise edges~\cite{wang2022ada}. When messages pass along these noise edges, the face features are polluted, degrading the performance of face clustering.
Ada-NETS~\cite{wang2022ada} was proposed to address the noise edges problem in GCN-based face clustering. 
However, our tests have revealed that the prediction of $k_{off}$ can be inaccurate, often deviating near the optimal $k$. This inaccuracy can lead to suboptimal graph construction and, consequently, lower clustering performance.

\subsection{Attention Mechanisms in Face Clustering}

Attention mechanisms~\cite{vaswani2017attention} have been increasingly incorporated into face clustering methods to address the issue of noise edges. Clusformer~\cite{nguyen2021clusformer} uses a self-attention mechanism to capture long-range dependencies between face nodes. By selectively attending to relevant nodes, it aims to reduce the impact of noise. LCE-PCENet~\cite{shin2023local} develops a local connectivity estimation network (LCENet) to excludes negative pairs from the KNN graph while maintaining sufficient positive pairs and a pairwise connectivity estimation network (PCENet) to determine whether or not two linked nodes belong to the same cluster. B-Attention~\cite{wang2022robust} is an upgraded version of Ada-NETS~\cite{wang2022ada}, but it is based on Transformer. However, Vanilla Transformer face challenges when applied to face clustering. Their tendency to overemphasize feature relationships may introduce noise, as they may assign high attention scores to irrelevant or noisy features~\cite{ye2024differential}. The Differential Transformer~\cite{ye2024differential}, in contrast, addresses this issue by using a differential attention mechanism. It calculates attention scores as the difference between two separate softmax attention maps, effectively canceling out noise attention. 
This method has shown great potential in language modeling, and we apply it to face clustering and proposed a Sparse Differential Transform (SDT) network that utilizes the inherent prior information in clustering to enhance the robustness of model.

\section{Methodology}

The framework of our face clustering method as shown in~\cref{fig:4}. Similar to existing methods~\cite{shin2023local, liu2024face}, we first construct a face graph. Subsequently, we employ a Transformer-based Adaptive Neighbor Discovery strategy to predict neighboring boundaries Top-K. Then we use the Top-K to refine the face graph. Due to the inherent inaccuracy of the predicted Top-K, noise nodes are inevitably retained. Therefore, we utilize a neural network (Sparse Differential Transformer) to alleviate the impact of noise nodes. Finally, we employ the Map Equation~\cite{rosvall2009map, liu2024face} method to perform face clustering.

\begin{figure*}[t]
	\centering
	\includegraphics[width=0.96\linewidth]{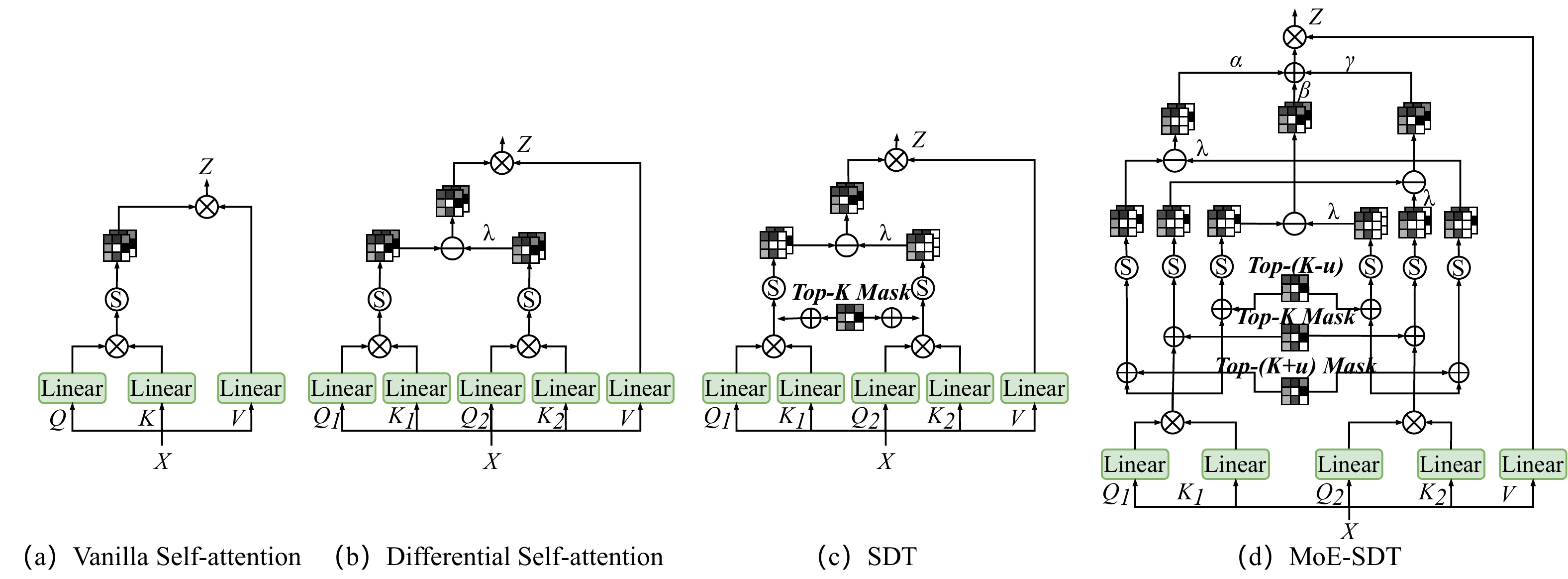}
	\vspace{-3mm}
	\caption{(a) Vanilla Self-attention. (b)Differential Self-attention. (c) Sparse Differential Self-attention (SDT). (d) The Mixture of Experts Sparse Differential Self-attention (MoE-SDT). } 
	\label{fig:3}
	\vspace{-3mm}
\end{figure*}

\begin{figure}[t]
	\centering
	\includegraphics[width=0.92\linewidth]{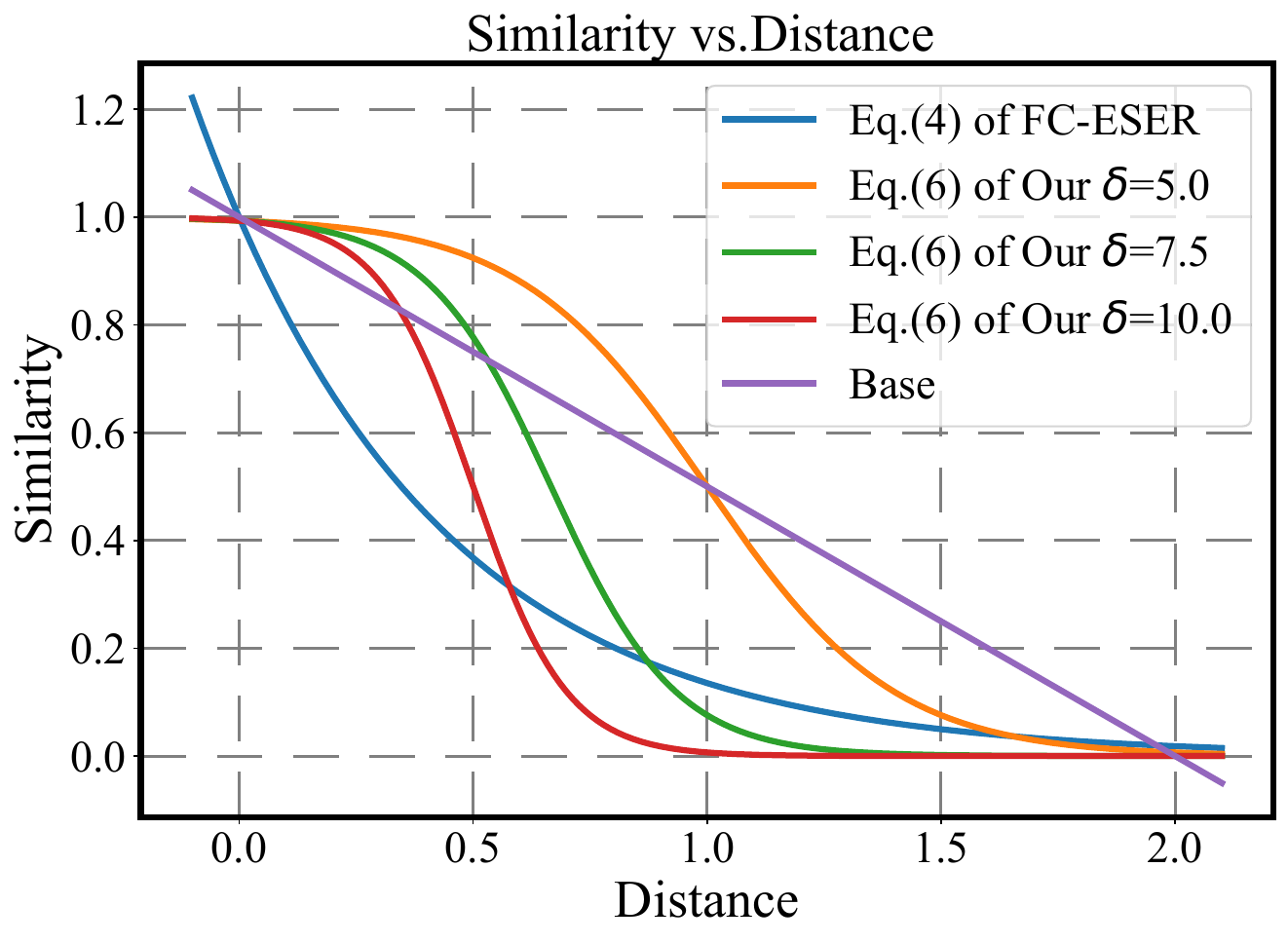}
	\vspace{-3mm}
	\caption{The distance transform function. } 
	\label{fig:5}
	\vspace{-3mm}
\end{figure}

\subsection{Top-K Jaccard Similarity Coefficient}

Given a set of $N$ face images, we extract face features $f_i \in \mathbb{R}^d$ for each image $i = 1,2,\cdots,N$, where $d$ is the dimension of the face features. For each sample $i$, we calculate the cosine similarity between $f_i$ and all other features $f_j$ as $a_{ij}=\frac{f_i\cdot f_j}{\|f_i\|\|f_j\|}$, and then select the $K$ nearest neighbors of $f_i$ based on the sorted cosine similarity values to construct the similarity matrix $\mathcal{S} \in \mathbb{R}^{N\times K}$ and adjacency matrix $\mathcal{N} \in \mathbb{R}^{N\times K}$. 
Simple distance metrics, such as cosine similarity, may not fully capture the complex relationships between faces. Inspired by Ada-NETS~\cite{wang2022ada} and FC-ESER~\cite{liu2024face}, we use the Jaccard similarity coefficient~\cite{zhong2017re} to measure the similarity between two samples in the face graph. Let $\mathcal{N}_i$ and $\mathcal{N}_j$ be the sets of $k$ nearest neighbors of two face samples $i$ and $j$ respectively. In Ada-NETS, the Jaccard similarity coefficient $J(\mathcal{N}_i, \mathcal{N}_j)$ between $i$ and $j$ is defined as:
\vspace{-1mm}
\begin{gather}
	J(\mathcal{N}_i, \mathcal{N}_j) = \frac{|\mathcal{N}_i \cap \mathcal{N}_j|}{|\mathcal{N}_i \cup \mathcal{N}_j|}.
\end{gather}
FC-ESER improves the Jaccard similarity coefficient and redefines it as neighbor-based edge probability $\widetilde{p}_{ij}$:
\begin{gather}
	\widetilde{p}_{ij} = J(\mathcal{N}_i, \mathcal{N}_j) = \frac{\sum_{h_{ij} \in {\mathcal{M}}_{ij}} (\hat{p}_{ih}+\hat{p}_{hj})}{\sum_{h_i \in {\mathcal{N}}_i} \hat{p}_{ih}+\sum_{h_j \in {\mathcal{N}}_j} \hat{p}_{hj}}. \label{eq:pijew}
\end{gather}
where ${\mathcal{M}}_{ij}={\mathcal{N}}_i \cap {\mathcal{N}}_j$ is the intersection of ${\mathcal{N}}_i$ and ${\mathcal{N}}_j$. $h_{ij}$ represents the common neighbors of samples $i$ and $j$. $h_i$ and $h_j$ represent neighbors of samples $i$ and $j$, respectively. $\hat{p}$ is the edge probability between $i$ and $j$ that defined in FC-ESER as:
\begin{minipage}{0.38\linewidth}
    \begin{equation}
      d_{ij} = 2 - 2 * a_{ij}.
    \end{equation}
\end{minipage}%
\begin{minipage}{0.33\linewidth}
    \begin{equation}
      p_{ij} = e^{-\frac{d_{ij}}{\tau}}. \label{eq:pije}
    \end{equation}
\end{minipage}%
\begin{minipage}{0.27\linewidth}
    \begin{equation}
      \hat{p}_{ij} = \frac {p_{ij}}{s_i}. \label{eq:pijeP}
    \end{equation}
\end{minipage}
where $a_{ij}$ is the cosine similarity between $f_i$ and $f_j$, $\tau$ is a temperature parameter. $s_i=\sum_{j \in {\mathcal{N}}_i} p_{ij}$ is the sum of edge probabilities $p_{ij}$ of sample $i$. 
However, \cref{eq:pije} has a critical flaw, it reduces the similarity between two face features, leading to two significant issues: 1) Faces belonging to the same identity are fragmented into multiple subclusters (reducing recall). 2) The differences in Jaccard similarity coefficients are diminished, causing different identities to be clustered together (reducing precision). In addition, Ada-NETS and FC-ESER introduce a large number of irrelevant samples when calculating the Jaccard similarity coefficient, which also results in the Jaccard similarity coefficients between different identity samples being very close.
To address first problem, we redefine the distance transform function \cref{eq:pije} as:
\begin{gather}
	p_{ij} = \frac{1}{1 + e^{\delta d_{ij} + \epsilon}} \label{eq:pijsi}
\end{gather}
where $\delta = 7.5$ and $\epsilon = -5$ in this paper, as show in~\cref{fig:5}. 
The sigmoid-like function used in our formula helps to amplify small differences in the original distance, making it easier to distinguish between similar and dissimilar face images. 
To address the second issue, we refine the Jaccard similarity formula (\cref{eq:pijew}) based on an adaptive neighbor discovery strategy. Specifically, we employ the adaptive filter of Ada-NETS, replace LSTM with Transformer, to predict the optimal number of neighbors, denoted as Top-K. When calculating the Jaccard similarity coefficient, we consider only the samples preceding Top-K to reduce the number of irrelevant samples. 
And to further improve noise robustness, we redefine the formula of Top-K as $\hat{k}$:
\begin{gather}
	\hat{k}=\left\{
	\begin{aligned}
		& 10 * \lceil\tfrac{\text{Top-K}}{10}\rceil, & \text{Top-K} < K \\
		& K, & \text{Top-K} \geq K
	\end{aligned}
	\right.
\end{gather}
Therefore, the neighbor-based edge probability $\widetilde{p}_{ij}$ is redefined as:
\begin{gather}
	\widetilde{p}_{ij} = \frac{\sum_{h_{ij} \in {\mathcal{M}}_{ij}^{\hat{k}_i}} (\hat{p}_{ih}+\hat{p}_{hj})}{\sum_{h_i \in {\mathcal{N}}_i^{\hat{k}_i}} \hat{p}_{ih}+\sum_{h_j \in {\mathcal{N}}_j^{\hat{k}_i}} \hat{p}_{hj}} \label{eq:pijsik}
\end{gather}
where $\hat{k}_i$ denotes the optimal number of neighbors of sample $i$. ${\mathcal{M}}_{ij}^{\hat{k}_i}={\mathcal{N}}_i^{\hat{k}_i} \cap {\mathcal{N}}_j^{\hat{k}_i}$ is the intersection of ${\mathcal{N}}_i$ and ${\mathcal{N}}_j$ that preceding $\hat{k}$. Through \cref{eq:pijsi} and \cref{eq:pijsik}, we can effectively improve the face clustering ability.

\begin{table*}[t]
	\centering
	\renewcommand\tabcolsep{8pt}
	\scalebox{0.92}{
	\begin{tabular}{l|cc|cc|cc|cc|cc}
		\hline
		\#unlabeled & \multicolumn{2}{c|}{584K} & \multicolumn{2}{c|}{1.74M} & 
		\multicolumn{2}{c|}{2.89M} & 
		\multicolumn{2}{c|}{4.05M} &\multicolumn{2}{c}{5.21M} \\ \cline{1-11}
		Method / Metrics  & $F_{P}$ & $F_{B}$ & $F_{P}$ & $F_{B}$  & $F_{P}$ & $F_{B}$ &$F_{P}$ & $F_{B}$ &$F_{P}$ & $F_{B}$ \\
		\hline\hline
		K-Means~\cite{lloyd1982least}&79.21&81.23&73.04& 75.20& 69.83& 72.34& 67.90& 70.57& 66.47& 69.42\\
		HAC~\cite{sibson1973slink}&70.63&70.46&54.40& 69.53& 11.08& 68.62& 1.40& 67.69& 0.37& 66.96\\
		DBSCAN~\cite{ester1996density}&67.93&67.17&63.41& 66.53& 52.50& 66.26& 45.24& 44.87& 44.94& 44.74\\
		ARO~\cite{otto2017clustering}&13.60&17.00&8.78& 12.42& 7.30& 10.96& 6.86& 10.50& 6.35& 10.01\\
		CDP~\cite{lloyd1982least} &75.02&78.70&70.75&75.82&69.51&74.58&68.62&73.62&68.06&72.92\\
		L-GCN~\cite{wang2019linkage} &78.68&84.37&75.83&81.61&74.29&80.11&73.70&79.33&72.99&78.60\\
		LTC~\cite{yang2019learning} &85.66&85.52&82.41&83.01&80.32&81.10&78.98&79.84&77.87&78.86\\
		GCN(V+E)~\cite{yang2020learning}&87.93&86.09&84.04&82.84&82.10&81.24&80.45&80.09&79.30&79.25\\
		Clusformer~\cite{nguyen2021clusformer}&88.20&87.17&84.60&84.05&82.79&82.30&81.03&80.51&79.91&79.95\\
		STAR-FC~\cite{shen2021structure}&91.97&89.96&88.28&86.26&86.17&84.13&84.70&82.63&83.46&81.47\\
		FaceT~\cite{ye2021learning}&91.12&90.50&89.07&86.84&86.78&85.09&84.10&84.67&83.86&83.86\\
		Pair-Cls~\cite{liu2021learn}&90.67&89.54&86.91&86.25&85.06&84.55&83.51&83.49&82.41&82.40\\
		Ada-NETS~\cite{wang2022ada}&92.79&91.40&89.33&87.98&87.50&86.03&85.40&84.48&83.99&83.28\\
		Chen et al.~\cite{chen2022mitigating}&93.22&92.18&90.51&89.43&89.09&88.00&87.93&86.92&86.94&86.06\\
		FaceMap~\cite{yu2022facemap}&94.27&92.55&91.31&89.67&89.32&88.20&87.74&87.11&86.37&86.29\\
		Wang et al.~\cite{wang2022robust}&94.94&93.67&91.74&90.81&89.50&89.15&87.04&87.81&85.40&86.76\\
		LCEPCE~\cite{shin2023local}&94.64&93.36&91.90&90.78&90.27&89.28&88.69&88.15&87.35&87.28\\
		CLIP-Cluster~\cite{shen2023clip}&94.22&-&91.44&89.44&89.95&87.75&88.93&86.78&87.99&85.85\\
		FC-ESER~\cite{liu2024face}&95.28&93.85&92.94&91.54&91.61&90.38&90.44&89.50&89.40&88.80\\
		\hline\hline
		\textbf{Diff-Cluster (Our)} &\textbf{95.46}&\textbf{94.14}&\textbf{93.14}&\textbf{91.87}&\textbf{91.73}&\textbf{90.71}&\textbf{90.89}&\textbf{89.78}&\textbf{90.08}&\textbf{89.14}\\
		\hline
	\end{tabular}}
	\vspace{-1mm}
	\label{tab:exp_ms1m}
	\caption{Comparison on face clustering when training with 0.5M face images and testing with different numbers of unlabeled face images. All results are obtained on the MS1M dataset. The proposed SDT consistently outperforms other face clustering baselines on different scale of testing data.}
	\vspace{-2mm}
	\label{tab:table1}
\end{table*}

\subsection{Sparse Differential Transformer}
The predicted Top-K near its optimal number of neighbors, which may either omit relevant nodes or introduce irrelevant ones. This necessitates a thorough examination of the relationships between the central node and its neighboring nodes near the Top-K. To address this problem, we develop a Transformer-based model to predict these relationships in this paper.

However, Vanilla Transformer, when applied to face clustering, often face the problem of overemphasizing feature relationships, which can introduce noise and disrupt the clustering process. To address this issue, we propose the Sparse Differential Transformer (SDT), as shown in~\cref{fig:3} (a). Our SDT builds on the Differential Transformer~\cite{ye2024differential}, which uses a differential attention mechanism to cancel attention noise. In the Differential Transformer, the attention scores are calculated as the difference between two separate softmax attention maps, as shown in~\cref{fig:3} (c). The formula for the Differential Transformer is given by: 
\begin{gather}
	[Q_1, Q_2] = xW_Q, [K_1, K_2] = xW_K, V = xW_V  \\
	F_{Att} = (S(\tfrac{Q_1K_1^T}{\sqrt{d}}) - \lambda S(\tfrac{Q_2K_2^T}{\sqrt{d}}))V
\end{gather}
where $x$ is the face feature, $Q_1, Q_2, K_1, K_2\in\mathbb{R}^{N \times \tfrac{d}{2}}$ and $V\in\mathbb{R}^{N \times d}$ denote the query, key and value. $W_Q, W_K, W_V\in\mathbb{R}^{d \times d}$ are the weight of linear projection model. $F_{Att}$ denotes the output of the DIFF Attention, $S(\cdot)$ is the $softmax(\cdot)$ function. $\lambda$ is a learnable scalar:
\begin{gather}
	\lambda = exp(\lambda_{q1} * \lambda_{k1}) - exp(\lambda_{q2} * \lambda_{k2}) + \lambda_{init}
\end{gather}
where $\lambda_{q1}, \lambda_{q2}, \lambda_{k1}, \lambda_{k2} \in\mathbb{R}^{\frac{d}{2}}$ are learnable parameters, and $\lambda_{init} = 0.8$ in this paper.

In a standard Transformer, attention weights are computed for all pairs of nodes, leading to a large number of interactions that may not be relevant. Therefore, we further enhance this mechanism by introducing sparsity. Our SDT model applies a sparsity constraint to the attention mechanism, ensuring that only the most relevant nodes are attended to, while irrelevant or noisy relationships are ignored.
The sparsification process is implemented through Top-k selection, where we utilize the predicted Top-K as the threshold. At each layer of the Transformer, we calculate the attention weights between nodes and retain only the most important relationships. These selected relationships are then passed through subsequent layers of the model, while the discarded relationships are effectively ignored. This reduces the model's sensitivity to noise, which is particularly useful in clustering tasks where irrelevant features can distort the clustering decision. The formula for the Sparse Differential Transformer is given by: 
\begin{gather}
	F_{Att} = (S(M(\tfrac{Q_1K_1^T}{\sqrt{d}})) - \lambda S(M(\tfrac{Q_2K_2^T}{\sqrt{d}})))V
\end{gather}
where $M$ denotes the Top-K Attention Mask. As previously discussed, Top-K frequently deviates from the true optimal value. It can result in relevant nodes being overlooked. To address this issue, we propose a Mixture of Experts Sparse Differential Self-attention (MoE-SDT), as shown in~\cref{fig:3} (d). We introduce attention mask before and after Top-K and incorporate learnable parameters to form a MoE system, thereby enhancing the model's robustness. The MoE-SDT can be formulated as:
\begin{equation}
	\begin{aligned}
	F_{Att} = & \alpha (S(M_{K-u}(\tfrac{Q_1K_1^T}{\sqrt{d}})) - \lambda S(M_{K-u}(\tfrac{Q_2K_2^T}{\sqrt{d}})))V + \\
	          & \beta (S(M_{K}(\tfrac{Q_1K_1^T}{\sqrt{d}})) - \lambda S(M_{K}(\tfrac{Q_2K_2^T}{\sqrt{d}})))V + \\
	          & \gamma (S(M_{K+u}(\tfrac{Q_1K_1^T}{\sqrt{d}})) - \lambda S(M_{K+u}(\tfrac{Q_2K_2^T}{\sqrt{d}})))V
	\end{aligned}
\end{equation}
where $M_{K-u}$, $M_{K-u}$ and $M_{K-u}$ denote the Top-(K-u), Top-K and Top-(K+u) Attention Mask, $u=5$ in this paper. $\alpha$, $\beta$, and $\gamma$ are the learnable parameters of MoE. 

\section{Experiments}

\begin{table*}[t]
	\small
	\addtolength{\tabcolsep}{1.0pt}
	\centering
	\begin{tabular}{ccccccccc}
		\hline
		Method & $\delta$ & $P_{P}$ & $R_{P}$ & $F_{P}$ & $P_{B}$ & $R_{B}$ & $F_{B}$ & $NMI$ \\
		\hline	 
		FC-ES + \cref{eq:pije} & - & 96.23 & 92.80 & 94.49 & 95.86 & 90.34 & 93.02 & 0.9815 \\ \hline
		\multirow{3}{*}{FC-ES + \cref{eq:pijsi}}  & 5.0 & 97.60 & 90.82 & 94.09 & 97.56 & 87.56 & 92.29 & 0.9795  \\
		& 7.5 & 96.58 & 92.84 & 94.67 & 95.98& 90.54 & 93.18 & 0.9820 \\ 
		& 10.0 & 95.40 & 93.51 & 94.45 & 94.20 & 91.54 & 92.85 & 0.9809 \\ \hline
		FC-ES + \cref{eq:pijsi} + \cref{eq:pijsik} (Top-K) & 7.5 & 97.05 & 93.07 & 95.02 & 96.34 & 90.78 & 93.48 & 0.9820 \\ \hline
	\end{tabular} 
	\vspace{-1mm}
	\caption{Comparison results under different distance metric function. $P_{*}$ and $R_{*}$ denote the precision and recall, respectively.
	}
	\label{tab:table2}
	\vspace{-1mm}
\end{table*}

\begin{figure*}[t]
	\centering
	\includegraphics[width=0.95\linewidth]{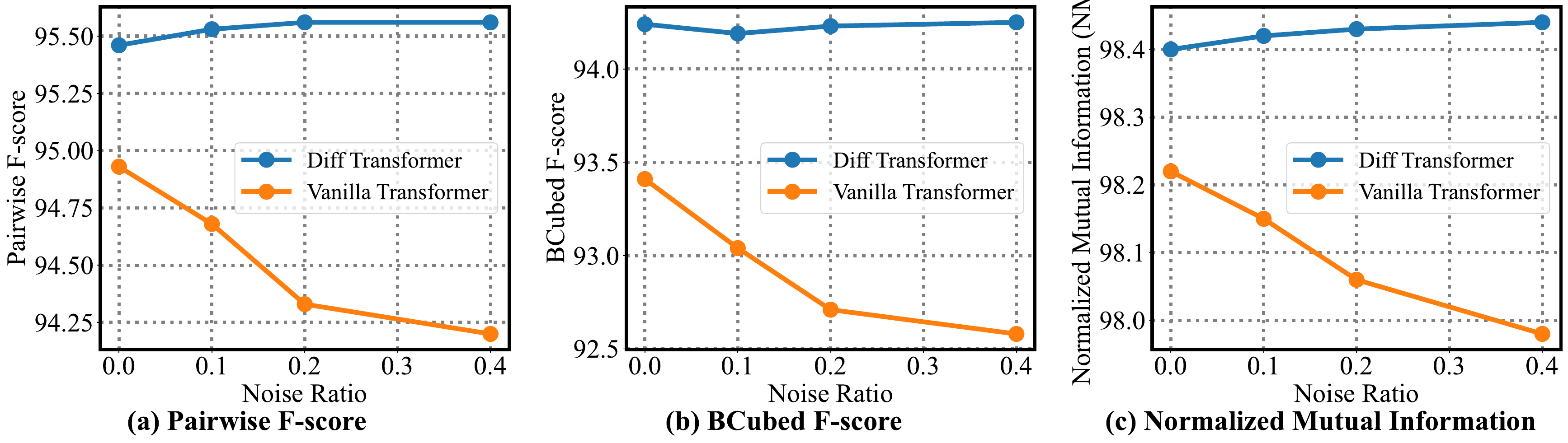}
	\vspace{-2mm}
	\caption{The performance of the Diff Transformer and Vanilla Transformer under different noise ratios.} 
	\label{fig:7-1}
	\vspace{-1mm}
\end{figure*}

\subsection{Experimental Settings}
\noindent\textbf{Datasets.} We conduct experiments on multiple datasets to thoroughly assess the effectiveness and generalization of our method. MS-Celeb-1M~\cite{guo2016ms, deng2019arcface} serves as the primary benchmark, containing 5.8M images of 86K identities after cleaning. Following GCN(V+E)~\cite{yang2020learning} and Ada-NETS~\cite{wang2022ada}, we train on part 0 and evaluate on parts 1, 3, 5, 7, and 9. To further validate generalization beyond the face domain, we also test on DeepFashion~\cite{liu2016deepfashion} and MSMT17~\cite{wei2018person}, representing clothes retrieval and person re-identification tasks, respectively.

\noindent\textbf{Metrics.} To evaluate clustering performance, we adopt Pairwise F-score ($F_{P}$)~\cite{shi2018face}, BCubed F-score ($F_{B}$)~\cite{bagga1998algorithms}, and Normalized Mutual Information ($NMI$)~\cite{strehl2002cluster}, which jointly measure precision and recall to assess how accurately the method groups faces into correct identities.

\noindent\textbf{Implementation Details.} To construct the face graph, we set K=80 for MS1M and K=40 for MSMT17, and the neural network-based relationship predictor model score threshold $\eta$ is 0.90 for MS1M and 0.88 for MSMT17. For the SDT, we adopt the architecture similar to the original Differential Transformer, with modifications to incorporate the sparsity constraint. The number of layers is set to 3, the number of attention heads is set to 8, and the hidden dimension is 1024. The learning rate for training the overall model is set to 1e-2, and we use the SGD optimizer with a momentum 0.9 and weight decay of 1e-4.

\begin{table}[t]
	\small
	\addtolength{\tabcolsep}{1.3pt}
	\centering
	\begin{tabular}{cccc}
		\hline
		Method & $F_{P}$ & $F_{B}$ & $NMI$  \\
		\hline	 
		Vanilla Transformer  & 94.25 & 92.73 & 0.9805  \\
		Vanilla Transformer + Top-K  & 94.78 & 93.25 & 0.9818  \\
		Diff Transformer  & 95.05 & 93.59 & 0.9830  \\
		Diff Transformer + Top-K  & 95.34 & 93.93 & 0.9839  \\
		Diff Transformer + MoE Top-K  & 95.46 & 94.14 & 0.9840  \\
		\hline
	\end{tabular} 
	\vspace{-1mm}
	\caption{Comparison results under different Transformer method with or without Top-K attention mask. Diff Transformer + Top-K means SDT and Diff Transformer + MoE Top-K means MoE-SDT.}
	\label{tab:table3}
	\vspace{-1mm}
\end{table}

\subsection{Comparison with State-of-the-Art Methods}
\cref{tab:table1} presents a comprehensive comparison of face clustering performance on the MS1M dataset. Our method achieves SOTA results across all tested scales of data (584K to 5.21M), highlighting the effectiveness of the proposed Sparse Differential Transformer (SDT). Traditional clustering algorithms such as K-Means~\cite{lloyd1982least}, HAC~\cite{sibson1973slink}, and DBSCAN~\cite{ester1996density} exhibit significant performance degradation as the dataset scale increases. GCN-based methods like L-GCN~\cite{wang2019linkage} and GCN(V+E)~\cite{yang2020learning} show improved robustness but still suffer from noise edges caused by fixed $k$NN graphs, which propagate erroneous features during message passing. Our adaptive neighbor discovery strategy alleviates this issue, maintaining high performance with only a minor drop (from $F_P=95.46$ at 584K to $90.08$ at 5.21M). Transformer-based methods such as Clusformer~\cite{nguyen2021clusformer} and LCEPCE~\cite{shin2023local} better capture long-range dependencies but are prone to attention noise. In contrast, our SDT employs a Top-K sparse differential attention mechanism that selectively emphasizes relevant nodes while masking irrelevant ones, effectively reducing noise interference.

\subsection{Ablation Study}

\subsubsection{Analysis of Noise Robustness in Face Clustering}
The Figure~\ref{fig:7-1} illustrates the performance of the Diff Transformer and Vanilla Transformer under different noise ratios (10\%, 20\%, and 40\%) on the MS1M 584K dataset, measured by Pairwise F-score ($F_{P}$), BCubed F-score ($F_{B}$) and Normalized Mutual Information ($NMI$) respectively. Specifically, in the similarity matrix, we randomly select 10\% , 20\% , and 40\% cosine similarity, respectively, and randomly increase their values from 0 to 1 and normalize them between 0 and 1.
In Figure~\ref{fig:7-1}, as the noise ratio increases, the $F_{P}$, $F_{B}$ and $NMI$ of the Diff Transformer shows a slight upward trend, while the Vanilla Transformer's performance decreases significantly. This indicates that the Diff Transformer is more robust to noise and can maintain and even slightly improve clustering performance with increased noise. The reason may be that the random noise addition inadvertently increases the similarity of faces from the same identity that were previously under-recognized due to factors like pose, expression, illumination, and age variations, making it easier to cluster them together.

\subsubsection{Impact of Top-K Jaccard Similarity Coefficient}
\cref{tab:table2} evaluates the impact of different distance metrics on clustering performance. Our proposed sigmoid-based metric \cref{eq:pijsi} outperforms the baseline (\cref{eq:pije} of FC-ES~\cite{liu2024face}) by 0.18\% in $F_P$ and 0.16\% in $F_B$. The parameter $\delta$ (controls the steepness of the sigmoid curve) significantly affects performance. 
It amplifies subtle differences in feature distances, particularly near decision boundaries. 
This amplification reduces fragmentation of same-identity clusters and prevents merging of distinct identities. 
The optimal $\delta=7.5$ balances sensitivity and specificity, avoiding over-amplification (which introduces false positives at $\delta=10.0$) or under-amplification (which increases false negatives at $\delta=5.0$). In addition, when calculating the Jaccard similarity coefficient, we consider only the samples preceding Top-K to reduce the number of irrelevant samples.

\subsubsection{Diff Transformer vs. Vanilla Transformer}
\cref{tab:table3} compares the Vanilla Transformer and the proposed Diff Transformer, both with and without Sparse Attention. While effective at capturing long-range dependencies and modeling complex relationships, the Vanilla Transformer tends to overemphasize irrelevant or noisy features, degrading clustering performance.
Our proposed Diff Transformer addresses this limitation by introducing a differential attention mechanism that effectively cancels out attention noise. We also introduce Sparse Attention mechanism further improves the performance of the Diff Transformer. Our Sparse Attention mechanism focuses on the most relevant nodes by applying a Top-K selection strategy, which retains only the most important relationships and masks the attention of unrelated nodes. This sparsity constraint reduces the model's sensitivity to noise and enhances its robustness.

\begin{table}[t]
	\small
	\addtolength{\tabcolsep}{2.2pt}
	\centering
	\begin{tabular}{cccccc}
		\hline
		Method & $K$ & $\eta$ & $F_{P}$ & $F_{B}$ & $NMI$\\
		\hline	 
		\multirow{7}{*}{MoE-SDT}  & 80 & 0.5 & 95.34 & 94.02 & 0.9836 \\
		& 80 & 0.6 & 95.37 & 94.03 & 0.9837 \\ 
		& 80 & 0.7 & 95.38 & 94.06 & 0.9837 \\ 
		& 80 & 0.8 & 95.40 & 94.07 & 0.9838 \\ 
		& 80 & 0.9 & 95.46 & 94.14 & 0.9840 \\ \cline{2-6}
		& 40 & 0.9 & 95.32 & 93.92 & 0.9839 \\ 
		& 160 & 0.9 & 95.34 & 94.02 & 0.9836 \\ \hline
	\end{tabular} 
	\vspace{-1mm}
	\caption{Comparison results with different $k$NN face graph and the neural network-based relationship predictor score threshold $\eta$.
	}
	\label{tab:table6}
	\vspace{-1mm}
\end{table}

\begin{table}[t]
	\small
	\addtolength{\tabcolsep}{1.3pt}
	\centering
	\begin{tabular}{ccc}
		\hline
		Method & $F_{P}$ & $F_{B}$  \\
		\hline	 
		GCN(V+E)~\cite{yang2020learning}&50.27&64.56 \\
		STAR-FC~\cite{shen2021structure}&58.80&66.92 \\
		Ada-NETS~\cite{wang2022ada}&64.39&73.39 \\
		Chen et al.~\cite{chen2022mitigating}&58.59&68.24 \\
		FaceMap~\cite{yu2022facemap}&68.64&75.42 \\
		FC-ESER~\cite{liu2024face}&70.71&75.66 \\ \hline
		Our&70.83&75.73 \\	
		\hline
	\end{tabular} 
	\vspace{-1mm}
	\caption{Comparison results with the existing clustering methods on MSMT17 dataset.}
	\label{tab:table4}
	\vspace{-1mm}
\end{table}

\subsubsection{Impact of $k$NN Graph and Score Threshold $\eta$}
\cref{tab:table6} presents a comprehensive evaluation of the impact of different $k$NN graph configurations and score threshold $\eta$ on the performance of our proposed Sparse Differential Transformer (SDT) for face clustering. The results demonstrate that our proposed SDT is robust to variations in the score threshold $\eta$ and the $k$NN graph configuration. The SDT's ability to reduce the impact of noise nodes and adaptively discover neighbors leads to more accurate clustering outcomes and reduced sensitivity to parameter choices. This robustness is a significant advantage over traditional methods, which often struggle with noise edges and are highly sensitive to the choice of $\eta$ and $k$. 

\subsubsection{Robustness of Sparse Differential Transformer}
\cref{tab:table4} presents the results of the performance of our method on non-face datasets MSMT17. The results show that our method achieves superior performance on this dataset, indicating that our method can effectively handle different types of data beyond face images. 
In addition, \cref{tab:table5} presents the results of the robustness of our SDT by replacing the Vanilla Transformer in the LCEPCE~\cite{shin2023local}. The results show that our SDT significantly improves the performance of the LCEPCE method on both the MS1M and DeepFashion datasets. On the MS1M dataset, the LCEPCE with our SDT achieves $F_{P}=94.68$ and $F_{B}=93.31$, compared to $F_{P}=93.43$ and $F_{B}=92.44$ with the Vanilla Transformer. Our SDT addresses the noise attention problem in vanilla Transformers, enabling the model to focus more precisely on the most relevant features and ignore irrelevant or noisy relationships. These results also demonstrate the robustness and effectiveness of our SDT in improving the performance of existing face clustering methods.

\begin{table}[t]
	\small
	\addtolength{\tabcolsep}{2.2pt}
	\centering
	\begin{tabular}{cccc}
		\hline
		Dateset & Method & $F_{P}$ & $F_{B}$ \\
		\hline	 
		\multirow{2}{*}{MS1M 584K}  & LCEPCE + Vanilla$\dag$ & 93.43 & 92.44  \\
		& LCEPCE + Diff & 94.68 & 93.31 \\ \hline
		\multirow{2}{*}{DeepFashion}  & LCEPCE + Vanilla$\dag$ & 38.43 & 60.55  \\
		& LCEPCE + Diff & 41.76 & 63.93 \\ \hline
	\end{tabular} 
	\vspace{-1mm}
	\caption{Comparison results with Vanilla Transformer or Diff Transformer on LCEPCE method. $\dag$ represents the results of our replication based on their open-source code.
	}
	\label{tab:table5}
	\vspace{-1mm}
\end{table}

\section{Conclusion}
In this paper, we proposed a novel face clustering framework that enhances noise resilience and neighbor prediction accuracy in graph-based clustering. Traditional $k$NN-based methods often suffer from noise edges, leading to unreliable clusters. To address this, we introduced the Top-K Jaccard Similarity Coefficient to refine face similarity measurement and designed a neural predictor to adaptively estimate optimal neighbor relationships. Furthermore, the proposed Sparse Differential Transformer (SDT) employs a Top-K sparse attention mechanism to suppress irrelevant feature dependencies and strengthen anti-noise capability. Extensive experiments on MS1M, DeepFashion, and MSMT17 confirm that our approach consistently outperforms state-of-the-art methods in both accuracy and robustness. Overall, the integration of adaptive neighbor discovery, enhanced distance metrics, and noise-resilient attention enables a scalable and generalizable clustering solution across face and non-face domains.

\bibliography{aaai2026}

\end{document}